\documentclass[a4paper,10.5pt]{elsarticle}
\usepackage{geometry}
\usepackage{makecell}
\usepackage{extarrows}
\usepackage{subfigure}
\usepackage{mathrsfs}
\usepackage{lineno}
\usepackage{ulem}
\usepackage{graphicx}
\usepackage{subfigure}
\usepackage{fancyhdr}
\usepackage{amssymb}
\usepackage{multicol}
\usepackage{amsmath}
\usepackage[figuresright]{rotating}

\usepackage{hyperref}
\usepackage{cleveref}
\usepackage{multirow}

\begin{document}
\begin{frontmatter}

\title{Self-adaption grey DBSCAN clustering}

\author{Shizhan Lu$^{1*}$}

\address{ \small  $^1$College of Economics and Management, Nanjing University of Science and Technology, Nanjing,  210094,
 China (e-mail: lubolin2006@163.com) \\


}


\begin{abstract}
Clustering analysis, a classical issue in data mining,  is widely used in various research areas. This article aims at proposing a self-adaption grey DBSCAN clustering (SAG-DBSCAN) algorithm. First, the grey relational matrix is used to obtain the grey local density indicator, and then this indicator is applied to make self-adapting noise identification for obtaining a dense subset of clustering dataset, finally, the DBSCAN which automatically selects parameters is utilized to cluster the dense subset. Several frequently-used datasets were used to demonstrate the performance and effectiveness of the proposed clustering algorithm and to compare the results with those of other  state-of-the-art algorithms. The comprehensive comparisons indicate that our method has advantages over other compared methods.

\end{abstract}
\begin{keyword}
Clustering analysis,  density-based,  B-style grey relationship, SAG-DBSCAN.
\end{keyword}
\end{frontmatter}

\begin{multicols}{2}
\section{Introduction}

Cluster analysis, which focuses on the grouping and categorization of similar elements, is widely used in different  research areas,   such as climate predictions  \cite{KCP}, gene expression \cite{WTX},  bioinformatics \cite{PEA}, finance and economics \cite{HJD, LG},  and neuroscience \cite{GS, AHK}.

In general, different clustering methods can be basically classified as follows: density-based (DP \cite{RA}, DP-HD \cite{MR}, DBSCAN \cite{EM}, NQ-DBSCAN \cite{CHEN}   and  CSSub \cite{ZHY}); grid-based (CLIQUE \cite{AGR}, Gridwave \cite{DEC} and WaveCluster \cite{SG}); model-based (Gaussian parsimonious \cite{MUK}, Gaussian mixture models \cite{OHA}  and Latent tree models\cite{CT}); partitioning (K-means \cite{MJ, ZHR, ZHA}, K-partitioning \cite{CDW} and TLBO \cite{LK}); graph-based (SEGC \cite{WAJ}, ProClust \cite{PPI} and MCSSGC \cite{VIE});  and hierarchical (BIRCH \cite{ZHT}, K-d tree and Quadtree \cite{DAJ}  and CHAMELEON \cite{GAK}) approaches.

DBSCAN \cite{EM} is a representative density-based algorithm which  clusters data by defining the density criterion with two parameters, Eps-distance and MinPts. NQ-DBSCAN  \cite{CHEN}, AA-DBSCAN \cite{KIM}, RNN-DBSCAN \cite{BRY}, ReCon-DBSCAN \cite{ZHTM} and ReScale-DBSCAN \cite{ZHTM}  are some up-to-date developments of DBSCAN.  As a disadvantage, DBSCAN and its extensions are difficult for their parameters to be set, which  are ruleless  on account of the different densities for variant datasets.

Grey relational analysis is  a significant tool for data mining \cite{WUD, YHH} and clustering analysis\cite{CKC,  LCH, LIX}. In this article, grey relational analysis is applied to obtain grey local density indicator for every object in clustering dataset. And then, grey local density indicator is applied to make noise identification for obtaining a dense subset. The dense subset which  the border points of $i$th cluster are far away from the border points of $j$th cluster is easily to set parameters for DBSCAN. This method  overcomes the disadvantage of DBSCAN which is difficult for its parameters to be set for variant datasets.

The remainder of this article is organized as follows: Section 2 presents a grey local density indicator, proposes a method for self-adaption noise identification and proposes a self-adaption grey-DBSCAN clustering method. Section 3  demonstrates our algorithms by some numerical experiments of both simulated and real datasets and make comparisons with the state-of-the-art clustering algorithms. Section 4 gives a conclusion.

\section{Proposed methods}

The main framework of the SAG-DBSCAN algorithm can be described as follows. Step 1  obtains the matrix of the B-style grey relationship degree and grey local  density indicator $\rho$. Step 2  utilizes linear regression to obtain dense subset $C$ (as shown in Fig. 1 (b)). Step 3  applies DBSCAN algorithm to cluster dense subset (as shown in Fig. 1 (c)). Step 4  assigns the object in $X-C$ to its nearest cluster (as shown in Fig. 1 (d)).

\begin{figure*}[hbt]
\centerline{\includegraphics[scale=0.46]{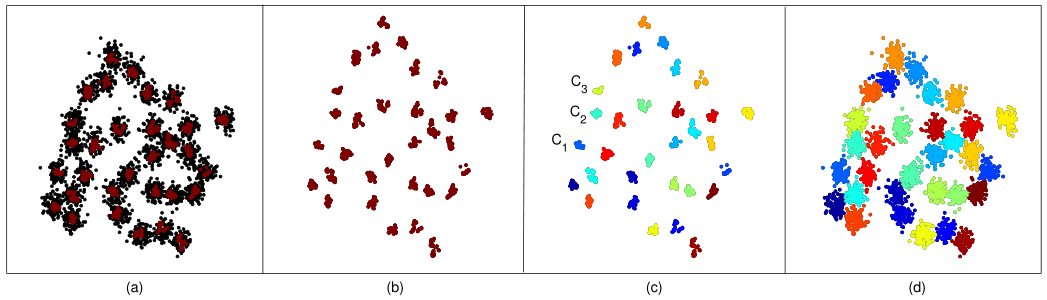}}
\caption{The main framework of the SAG-DBSCAN algorithm}
\label{}
\end{figure*}

\subsection{Grey local density indicator}

For all $x_i=(x_i(1),x_i(2),\cdots,x_i(N))$ and $x_j=(x_j(1),x_j(2),\cdots,x_j(N))$ in N-dimensional dataset $X$,

\begin{equation}
\gamma(x_i,x_j)=\frac{1}{1+d^{(0)}_{ij}/N+d^{(1)}_{ij}/(N-1)+d^{(2)}_{ij}/(N-2)}
\end{equation}
is the B-style grey relationship degree of $x_i$ and $x_j$ \cite{DJL, FEW}, where $d^{(0)}_{ij}=\sum\limits_{k=1}^N|x_i(k)-x_j(k)|$, $d^{(1)}_{ij}=\sum\limits_{k=1}^{N-1}|x_i(k+1)-x_j(k+1)-x_i(k)+x_j(k)|$ and  $d^{(2)}_{ij}=\sum\limits_{k=2}^{N-1}|x_i(k+1)-x_j(k+1)-2(x_i(k)-x_j(k))+x_i(k-1)-x_j(k-1)|$. $G=[\gamma(x_i,x_j)]_{n\times n}$ is denoted  as  B-style grey relationship degree matrix.

B-style grey relationship degree is  superior to other grey relationship degrees and Euclidean distance function for describing the objects' relationships  about object displacement, such as the DrivFace dataset in our experiments. Meanwhile, it can work well for other simulated data (Euclidean space points) like Euclidean distance function. B-style grey relationship degree has strong  applicability and generality for various datasets, hence, it is  selected  to make following analysis.

KNN-density  is a frequently-used indicator to describe the local density indicator $\rho_i$ \cite{CYW, DUM}. A relatively straightforward and useful grey KNN-density indicator is projected as equation (2), where $|GKNN(x_i)|=k$ and $G(i,j)\geqslant G(i,t)$ for all $x_j\in GKNN(x_i)$ and $x_t\in X-GKNN(x_i)$.

\begin{equation}
\rho_i=\sum\limits_{x_j\in GKNN(x_i)} G(i,j),
\end{equation}

In general, the dense family of a cluster is composed of many objects with large grey relationship degree between each other, on the contrary, the border objects has small grey relationship degree with its neighbors. As shown in Fig. 1 (c), the dense subset $C$ is composed of many dense families $C_i$ ($1\leqslant i \leqslant t$). The object $x_j$ of dense family has a large value of $\rho_j$ via calculating by equation (2) if we set $k\leqslant min\{|C_i|: i=1,2,\cdots, t\}$, where $C_i$ is the dense family of  $i$th cluster.

\subsection{Self-adaption method for noise identification}

Data preprocessing is necessary for the grey local density indicator $\rho$ before the noise identification.  Let $\rho'=\{\rho_i': 1\leqslant i\leqslant n\}$ be the descending sequence of $\rho=\{\rho_j: 1\leqslant j\leqslant n\}$ and $V=\{v_i: 5\leqslant i\leqslant n\}$ be the mean smoothing sequence of $\rho'$, where $v_i=(\rho_i'+\rho_{i-1}'+\rho_{i-2}'+\rho_{i-3}'+\rho_{i-4}')/5$.\\

\begin{figure*}
\centering
\includegraphics[scale=0.46]{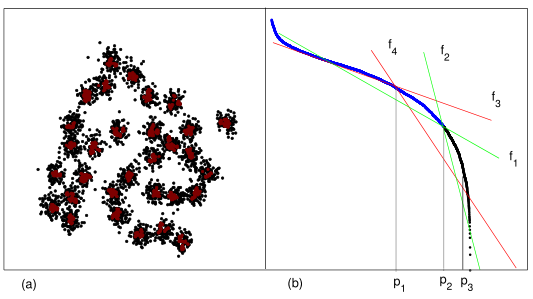}
\caption{The distinctions of border points and dense area points showed in $V$}
\label{}
\end{figure*}

As shown in Fig. 2, the elements of $V$ have two distinctly different distribution trends. The elements in $V$ are divided into two parts  $V_{p}^-=\{v_i:5\leqslant i\leqslant p\}$ and $V_{p}^+=\{v_i:p<i\leqslant n\}$ for linear regression, and $f_1$ and $f_2$ are the regression equations for $V_{p}^-$ and $V_{p}^+$, respectively.

\begin{equation}
\left\{
\begin{aligned}
e_1(i)=f_1(i)-v_i,  5\leqslant i\leqslant p, \\
e_2(i)=f_2(i)-v_i,  p<i\leqslant n.
\end{aligned}
\right.
\end{equation}
$R=\sum\limits_{5\leqslant i\leqslant p} |e_1(i)|+\sum\limits_{p<i\leqslant n}|e_2(i)|$ is the regression residual.

As shown in Fig. 2, we pick $p_1$, $p_2$ and $p_3$ in different positions for linear regressions and obtain three regression residuals $R_1$, $R_2$ and $R_3$, where $f_1$ and $f_2$ are the regression straight line with respect to $V_{p_2}^-$ and $V_{p_2}^+$, respectively.  The results show that $R_2<R_1$ and $R_2<R_3$.

After investigating the characteristic of the grey local density indicator $\rho$ and the smoothing sequence $V$, we can make a sequence of linear regressions to obtain a residual sequence $R=\{R_i:5+5\leqslant i\leqslant n-5\}$ (5 points for regression). If $R_p=min(R)$, the object $x_j$ is considered as a member of dense subset for the case $\rho_j\geqslant \rho'_p$. Then, the dense subset (as shown in Fig. 1 (c)) is obtained for clustering by DBSCAN method.

\subsection{Self-adaption grey DBSCAN clustering method (SAG-DBSCAN)}

The detailed processes of the SAG-DBSCAN algorithm are shown as Algorithm 1.

\rule{8cm}{0.5mm}

{\bf Algorithm 1:} SAG-DBSCAN algorithm.

\rule{8cm}{0.25mm}

{\bf Input:} Dataset, parameters $k\in N^+$ and $m\in N^+$.

{\bf Output:} The clustering result.

1. Use equation (1) to obtain B-style grey relationship degree matrix $G$.

2. Apply equation (2) to obtain grey local density indicator $\rho$ with respect to $k=|GKNN(x_i)|$.

3. Sort $\rho$ to obtain $\rho'$, smooth $\rho'$ to obtain $V$.

4. Obtain dense subset $C\subset X$ by  linear regression of $V$.

5. Set MinPts=$m$ and Eps-distance=$max\{d(x_i,x_i^{(m)}): x_i\in C\}$, where $x_i^{(m)}$ is the $m$th neighbor of $x_i$.

6. Apply DBSCAN to cluster dense subset $C$ with MinPts=$m$ and Eps-distance.

7. Assign the objects of $X-C$ to their nearest clusters.

\rule{8cm}{0.25mm}

The objects of $X-C$ can be assigned as follows: let $A\subset X$ be the subset that contains the already classified points and  $U\subset X$ be the subset of unclassified points. If $f(x_i',x_j')=min\{f(x_i,x_j):x_i\in A, x_j\in U\}$, then $x_j'$ is assigned to the category that contains $x_i'$.

The time complexities are  $\mathscr{O}(n^2)$ and $\mathscr{O}(n)$ for obtaining  B-style grey relationship degree matrix $G$ and dense subset $C$, respectively. The time complexity of DBSCAN is $\mathscr{O}(n^2)$ at the worst case.   Moreover, if $|C|=n'$, then $|X-C|=n-n'$, and the time complexity of assigning border points is hence $\mathscr{O}(n-n')$.

\section{Experiments}

In this section, we evaluate the performance and effectiveness of the proposed method on both simulation data and real data, and then compare it with some up to date methods: the affinity propagation algorithm (AP) \cite{FBJ}, automatic find of density peaks (ADPC) \cite{TLI}, Neighbor Query DBSCAB (NQ-DBSCAN) \cite{CHEN} and NK hybrid genetic algorithm (NKGA) \cite{TR}.

\subsection{Descriptions of Experiment data}

\begin{center}
\begin{table*}
\scriptsize
\label{tab:parametervalues}
\setlength{\tabcolsep}{3pt}
\begin{tabular}{|p{32pt}|p{28pt} p{25pt} p{26pt}|p{30pt}|p{28pt} p{25pt} p{26pt}|p{223pt}|}
\hline
Dataset& Instances &Features & Clusters   & Dataset& Instances &Features & Clusters  & Detail  \\
\hline
D31 &  3100 & 2 &31 & Iris &   150 & 4 &3 &three kinds of irises: Setosa, Versicolour and Virginica. Each kind has 50 samples \\
\hline
S1 &  5000 & 2 & 15  & Wifi &  2000 &7 &4  & 2000 times of signal records in 4 rooms, 500 records in each room  \\
\hline
R15 &  600 &2  &15   & Vertebral &  310 & 6 & 2  & 310 orthopaedic samples, 210 abnormal samples and 100 normal samples \\
\hline
Dim2 &  1350 &2 &9   & TumTyp & 801 &20531 & 5   & gene expressions of patients having different types of tumor: BRCA, KIRC, COAD, LUAD and PRAD.   \\
\hline
ShapeT & 10000 & 2 & 3 & DrivFace & 606 & 6400 & 4 &  dataset contains images sequences of subjects while driving in real scenarios. It is composed of 606
samples and acquired over different days from 4 drivers with several facial features.  \\
\hline
\end{tabular}
\caption{The simple description of datasets}
\label{tab1}
\end{table*}
\end{center}

First, some frequently-used datasets  obtained from different references are used to test the algorithms, such as R15 \cite{VCJ}, D31 \cite{VCJ},   S1 \cite{FP} and Dim2 \cite{FP1}   etc. And then a dataset, ShapeT (Fig. 3),  is constructed for the supplementary tests. All the simulation data are points of two-dimensional Euclidean space.

\begin{figure*}
\centering
\includegraphics[scale=0.46]{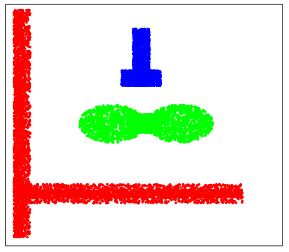}
\caption{ ShapeT}
\label{}
\end{figure*}

Several real-world datasets are used to test the performance of the proposed method, including a plant dataset: Iris \cite{FRA, HF}; a wireless signal dataset: Wifi \cite{RJG};  a human vertebral column dataset: Vertebral  \cite{BEE}; a  gene dataset: TumTyp \cite{WJN}; and a face image dataset: DrivFace \cite{KDC}.  Datasets   were taken from the UCI\footnote{http://archive.ics.uci.edu/ml/datasets.php} repository. Simple descriptions of these real datasets are provided in Table 1.

\subsection{Results and Comparisons}

\begin{table*}[htb]
\scriptsize
\centering
\label{tab:parametervalues}
\setlength{\tabcolsep}{3pt}
\begin{tabular}{|p{33pt}|p{10pt}p{20pt}p{20pt}p{50pt} p{60pt}|p{33pt}|p{10pt}p{20pt}p{20pt}p{50pt} p{60pt}|}
\hline
Dataset &  \multicolumn{5}{|c|}{The estimated number of clusters} &Dataset &  \multicolumn{5}{|c|}{The estimated number of clusters} \\
&  AP & ADPC   & NKGA &NQ-DBSCAN &SAG-DBSCAN & &  AP & ADPC & NKGA  &NQ-DBSCAN &SAG-DBSCAN   \\
\hline
D31 & 8  & {\bf 31 }  & 19 & {\bf 31}& {\bf 31} &Iris &   2  & 2 & 11  &{\bf 3} & {\bf 3}   \\
S1 & {\bf  15} &{\bf  15}  &14 &{\bf  15} &{\bf  15} & Wifi &  5 & {\bf 4}&1 &{\bf 4} &{\bf 4}  \\
R15 & 5 & {\bf 15} &{\bf 15}& {\bf 15}&{\bf 15} & Vertebral & 1 & 1&{\bf 2} & {\bf 2} &{\bf 2} \\
Dim2 &  7 &{\bf  9} &{\bf 9 }&{\bf 9 } &{\bf 9 } & TumTyp &  3 & {\bf 5} &6 &{\bf 5} &{\bf 5}  \\
ShapeT & 27 & 5  & 11 &{\bf 3}& {\bf 3} & DrivFace & 5 &6 &3  &5 &{\bf 4}\\
\hline
\end{tabular}
\caption{Number of clusters estimated by various methods}
\label{tab1}
\end{table*}

\begin{table*}[htb]
\scriptsize
\centering
\label{tab:parametervalues}
\setlength{\tabcolsep}{3pt}
\begin{tabular}{|p{32pt}|p{30pt}|p{22pt}p{22pt}p{22pt}p{28pt} p{36pt}|p{36pt}|p{30pt}|p{22pt}p{22pt}p{22pt}p{36pt} p{36pt}|}
\hline
 Dataset&  Measures & AP & ADPC  &NKGA &NQ-DBSCAN &SAG-DBSCAN & Dataset &  Measures &   AP & ADPC  &NKGA &NQ-DBSCAN &SAG-DBSCAN  \\
\hline
 D31 &  Accuracy    &0.2210 &  {\bf  0.9677} &   0.3539   &   0.5416 &  {\bf  0.9677}&      Iris &Accuracy & 0.5333  &  0.6667 &   0.4533    &  0.7867 &  {\bf  0.9067} \\
     & F-Score  & 0.3466 &  {\bf  0.9679} &  0.4537   &   0.6937   &{\bf  0.9679}&           &F-Score  & 0.4329 &   0.5714 &   0.5883   &   0.8697 &   {\bf   0.9168}\\
    &  ARI & 0.1704  & {\bf  0.9352} &  0.3290    &  0.1240  & {\bf  0.9352}&               &ARI &  0.4120  &  0.5681 &   0.2681    &   0.6789  &  {\bf  0.7592}\\
    & NMI   & 0.4929  &  {\bf 0.9573} &   0.6498  &  0.2994  &  {\bf 0.9573}&           &NMI &   0.4509  &  0.7337  &  0.0138   &   0.7603 &   {\bf  0.8057}\\
\hline
S1 &  Accuracy  & 0.7642  &  0.9262 &   0.6992   &  0.9614 &  {\bf  0.9932}&        Wifi &Accuracy &  0.1405 &   0.8625 &   0.2500  &   0.7545  & {\bf  0.9355 }  \\
    & F-Score &  0.7907  &  0.9332 &   0.7315  &   0.9647  & {\bf  0.9934}&       &F-Score  &   0.1671  &  0.8859  &  0.1000  &   0.8561 & {\bf   0.9402}  \\
     &  ARI &  0.6518   & 0.8915  &  0.5685   &   0.9378 &  {\bf  0.9858}&       &ARI & 0.1948  &  0.8103  &  0.0000    &  0.6868  &  {\bf 0.8470} \\
     & NMI  &  0.8382  &  0.9450 &   0.7878   &  0.9695  & {\bf  0.9895}&     &NMI &   0.2646  &  0.8309  &  0.0078  &   0.6531  & {\bf  0.8635} \\
\hline
R15&  Accuracy   &   0.2217  & 0.9917 &   0.8983    & 0.8200 &  {\bf  0.9933} &     Vertebral &Accuracy &   0.6774 &   0.6774  &  0.6645    &   0.1516 & {\bf   0.7710 } \\
     & F-Score  &   0.3416  &  0.9918 &   0.9035  &  0.9011 &  {\bf  0.9935} &            &F-Score  &  0.4038 &   0.4038  &  0.3992    &   0.1437 &  {\bf  0.7976 } \\
    &  ARI &  0.2574  &  0.9817 &   0.7968   &  0.7667 &  {\bf  0.9857}&                &ARI &   0.0335  &  0.0304   &  0.0166     & 0.2381  & {\bf  0.2916}\\
    & NMI  &   0.5460  & 0.9864  &  0.8705   &   0.3609   & {\bf 0.9893}&                &NMI &   0.0000 &  0.0000 &   0.0145   &  0.1635  & {\bf  0.3129} \\
\hline
Dim2 &  Accuracy  &0.8259  & {\bf  1.0000 } &  0.9289    & {\bf  1.0000} &  {\bf  1.0000}&        TumTyp &Accuracy & 0.3483 &  {\bf 0.9975} &  0.3059 &   {\bf 0.9975}& {\bf 0.9975} \\
     & F-Score & 0.8482  & {\bf  1.0000}  &  0.9384   &  {\bf 1.0000}  & {\bf  1.0000}&             &F-Score  & 0.5092& {\bf  0.9976}&   0.2666&    {\bf  0.9976} & {\bf  0.9976} \\
    &  ARI & 0.7549  & {\bf  1.0000} &   0.8714    & {\bf  1.0000}  & {\bf  1.0000}&                  &ARI &  0.1445   & {\bf 0.9938}  & 0.0025 &  {\bf 0.9938} & {\bf 0.9938}  \\
    & NMI  &0.8782  & {\bf  1.0000}  &  0.9337    & {\bf  1.0000 }  & {\bf  1.0000}&                  &NMI & 0.2577 & {\bf   0.9898} & 0.0042 & {\bf   0.9898}& {\bf   0.9898}  \\
\hline
ShapeT &  Accuracy   &   0.2667&  0.8808&   0.2799 & {\bf 1.0000} & {\bf 1.0000}  &     DrivFace &Accuracy &  0.3053 &  0.6436  & 0.2871 &  0.5924  & {\bf 0.6485}\\
    & F-Score   &  0.3600&   0.9586&   0.3732 &  {\bf 1.0000} & {\bf 1.0000}&           &F-Score  &  0.3685  & {\bf 0.7489} &  0.3303  &  0.5614 & 0.6853\\
    &  ARI  & 0.0910&   0.7819&  0.0938   &  {\bf 1.0000} &  {\bf 1.0000}&                  &ARI &    0.2394 &   0.4684 & 0.0076 &  {\bf 0.4701}  & 0.3663 \\
   & NMI   & 0.0930&   0.6104 &   0.0882   &  {\bf 1.0000} & {\bf 1.0000}&               &NMI &   0.2727  & 0.4285 &  0.0138  & {\bf 0.5018} &  0.4604 \\
\hline
\end{tabular}
\caption{The results' comparison  for different methods}
\label{tab1}
\end{table*}

\subsubsection{Results presentation}

Table 2 presents the number of clusters estimated by different methods. Table 3 shows the clustering results when compared with other methods.

To evaluate and compare the performance of the clustering methods, we apply the evaluation metrics: Accuracy, F-Score,  Adjusted Rand Index (ARI) \cite{DUL} and Normalized Mutual Information (NMI) \cite{ABB} in our experiments to do a comprehensive evaluation.   The higher
the value, the better the clustering performance for all these measures.
Compared with the best results of other algorithms,   our method has  relative advantages of 0.0936,0.3938, 0.0535 and 0.1494 (TABLE 3) with respect to Accuracy, F-Score, ARI and NMI for the Vertebral dataset, respectively.

\begin{figure*}
\centering
\includegraphics[scale=0.46]{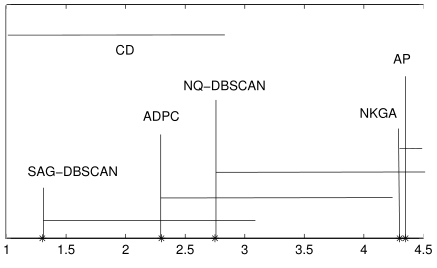}
\caption{Critical difference (CD) diagram of the post-hoc Nemenyi test ($\alpha$= 0.10)}
\label{}
\end{figure*}

We conduct the Friedman test with the post-hoc Nemenyi test \cite{ZHY} to examine whether the difference between any two
clustering algorithms is significant in terms of their average ranks. The
difference between two algorithms is significant if the gap between their ranks is
larger than CD. There is a line between two algorithms if the rank gap between
them is smaller than CD. This test shows that SAG-DBSCAN, ADPC and NQ-DBSCAN are significantly better than NKGA and AP. SAG-DBSCAN is the best performer of these algorithms, followed by ADPC (as shown in Fig. 4).

In summary, our method obtains better results with respect to the estimation of cluster number, Accuracy, F-Score, ARI and NMI, compared comprehensively with other methods.

\subsubsection{Parameter analysis} 

The parameters of the proposed method can be set with a reference of the number of objects in clustering dataset $X$. For these datasets, we set $m=3$ when $n<500$, $m=4$ when $500\leqslant n<1000$, $m=5$ when $1000\leqslant n<5000$ and $m=10$ when $n\geqslant 5000$. We set $k=ceil(2\%n)$ when $n<1000$, $k=ceil(1\%n)$ when $1000\leqslant n<2000$ and $k=20$ when $n\geqslant2000$.

The parameters of NQ-DBSCAN are shown in Table 4, for example Eps-distance=0.6 (left) and MinPts=23 (right) for D31 dataset. The parameter  settings of NQ-DBSCAN are random and ruleless for the datasets. It is thus  difficult to guess the right parameters for NQ-DBSCAN  if the results are unknown before clustering occurs.\\

{
\scriptsize   
\begin{tabular}{|p{30pt}p{30pt} p{30pt} p{35pt}p{35pt}|}
\hline
D31 & S1  & R15  & Dim2 & ShapeT  \\
0.6, 23 & 5000, 19  & 0.3, 6   & 5000, 10 & 0.2746, 10  \\
\hline
 Iris & Wifi & Vertebral & TumTyp & DrivFace  \\
 0.42, 5 & 6, 20 & 16, 7 & 166.05, 5 & 10.26, 5 \\
 \hline
\end{tabular}
}

{
\scriptsize
\ \ \ \ \ \ \ \ \ \text{Table 4: The parameter  settings of   NQ-DBSCAN}
}

The proposed method is robust. It can obtain the same clustering result when we select values for parameters $k$ and $m$ at wide intervals. For example, SAG-DBSCAN can obtain the same results for dataset Iris with $k\in[5,11]\cap N^+$ when $m=3$. For dataset Iris, NQ-DBSCAN cannot obtain the same results for three cases; Eps=0.41, Eps=0.42 and Eps=0.43, when MinPts=5.  NQ-DBSCAN is not robust in their parameters.

The  parameter of ADPC \cite{TLI} is set with $d_c=0.02$ for these datasets.  $d_c=0.02$ means that the parameter of ADPC takes the value at the position of first 2\% of all distances  \cite{TLI}. \\

\subsubsection{Comparisons and discussions}

AP \cite{FBJ}  is an unsupervised algorithm without any parameters. The parameters of  NKGA \cite{TR} are recommended  by the publication \cite{TR}. The algorithms of parameter-free or fixed parameter value may not be adaptive to various kinds of datasets.

ADPC can obtain good results in most instances. However, as a centroid-based method, ADPC and its variants cannot cluster
objects correctly when a category has more than one center, such as the ShapeT dataset which has  no
single point can be considered as the geometrical centroid of the T shape cluster.

The NQ-DBSCAN  produces good results for two-dimensional  data after it tunes  the parameters many times with  reference to two-dimensional figures. However, it does not work well for high-dimensional data, because these data cannot show well in two-dimensional figures. It is   hampered by the ruleless parameters when it deals with  multidimensional data. SAG-DBSCAN sets its parameters according to the number of objects, it is easier to set parameters than NQ-DBSCAN. SAG-DBSCAN obtains better results more easily than NQ-DBSCAN do, when faced with a new high-dimensional dataset that has no references to
known clustering results.

\section{Conclusion}

In this article, the SAG-DBSCAN algorithm is proposed, and then some simulation and real data are used to test the performance and effectiveness of the proposed method. Moreover, our proposed algorithm  is also compared with several frequently-used clustering algorithms, including the intelligent algorithm NKGA, the centroid-based algorithm ADPC,  the parameter-free self-adaption algorithm AP and an improved DBSCAN algorithm NQ-DBSCAN. The experiments indicate that our method obtains better results, in terms of the evaluation metrics (TABLE 3) and the estimated number of clusters (TABLE 2), than the other methods under comparison. Based on this work, it will be interesting to extend our method into a fully adaptive method in the future.




\end{multicols}
\end{document}